\documentclass[USenglish,twocolumn]{article}

\usepackage[utf8]{inputenc}

\newcommand{\affil}[1]{\date{{\small #1}}} \newcommand{\runningtitle}[1]{\pagestyle{myheadings} \markboth{#1}{#1}} \newcommand{\keywords}[1]{\newline {\bf Keywords:} #1} \newcommand{\articletype}[1]{} \newcommand{\journalname}[1]{} \newcommand{\startpage}[1]{}

\usepackage[hyphens]{url}
\usepackage{endnotes}
\let\footnote=\endnote

\begin{document}

\articletype{Research Article{\hfill}Open Access}

  \author{Bettina Berendt}

  \affil{KU Leuven, Department of Computer Science, Celestijnenlaan 200A, 3001 Heverlee, Belgium\\ 
E-mail: bettina.berendt@cs.kuleuven.be}

  \title{\huge AI for the Common Good?!\\
Pitfalls, challenges, and Ethics Pen-Testing\thanks{To appear in {\em Paladyn. Journal of Behavioral Robotics};
accepted on 27-10-2018}}

  \runningtitle{AI for the Common Good?!}

\journalname{Paladyn, J. Behav. Robot.}

\startpage{1}

\maketitle

  \begin{abstract}
Recently, many AI researchers and practitioners have embarked on research visions that involve doing AI for ``Good''.
This is part of a general drive towards infusing AI research and practice with ethical thinking.
One frequent theme in current ethical guidelines is the requirement that AI be good for all, or: contribute to the Common Good. But what is the Common Good, and is it enough to want to be good? Via four lead questions, I will illustrate challenges and pitfalls when determining, from an AI point of view, what the Common Good is and how it can be enhanced by AI. 
The questions are: What is the problem / What is a problem?, Who defines the problem?, What is the role of knowledge?, and What are important side effects and dynamics?
The illustration will use an example from the domain of ``AI for Social Good'', more specifically ``Data Science for Social Good''. 
Even if the importance of these questions may be known at an abstract level, they do not get asked sufficiently in practice, as shown by an exploratory study of 99 contributions
to recent conferences in the field.
Turning these challenges and pitfalls into a positive recommendation, as a conclusion I will draw on another characteristic of computer-science thinking and practice to make these impediments visible and attenuate them: ``attacks'' as a method for improving design. This results in the proposal of {\em ethics pen-testing} as a method for helping AI designs to better contribute to the Common Good.
\end{abstract}

\section{Introduction}

Artificial Intelligence (AI) is currently experiencing another ``summer'' in terms of perceived promises and economic growth.
At the same time, there are widespread debates around AI's perceived risks and negative impacts.
In response to the latter, AI researchers and practitioners are paying increasing attention to existing ethics codes, and they are drafting new ones.
In addition, many have embarked on research programs that explore how to do AI ``for Good''.
These two reactions are linked, at a high level, by the understanding that the goal of ethics codes is to encourage and ensure
``ethical'' professional conduct in the sense of this conduct being ``morally good or correct'' and ``avoiding activities [...] that do harm to people or the environment''.%
\footnote{\url{https://en.oxforddictionaries.com/definition/ethical}.
The term is often used, also in AI publications, in this everyday meaning rather 
than in the scientific meaning of relating to ethics.}
\footnote{All URLs referenced in this article were last retrieved on August 16, 2018.}

In addition to the goal to do ``good'', many current ethics codes and discussions go further and require that AI contribute to the Common Good.
 This term is not uniquely (and in many publications not at all) 
defined, but can be understood as the aim to be good for all.

The purpose of the current article is to 
investigate more closely the notion of AI for the Common Good by drawing on a wider literature, 
and to start a deeper discussion in the AI community about this goal and the way towards it. Towards this purpose,
I invite researchers and practitioners to ask four reflective questions of their research practices and projects. 
These questions can be used as {\em provocations}: 
interruptions of the flow of everyday practices designed to
``initiate critical reflection [...] on issues that are often otherwise overlooked, obscured or accepted as naturalised practice'' \cite[p. 225]{doi:10.1080/13645579.2016.1161346},
see \cite{boydCrawford} for the use of provocations to encourage reflection on big data.

The article is structured as follows.

The Common Good is a notion predating AI.
I will start from the definitions given in various AI ethics codes of the Common Good and related notions, 
and draw on selected discussions in political philosophy for deriving questions about these definitions and their operationalization for AI.
These definitions and questions are the subject of Section \ref{sec-CG-and-terms}.

Section \ref{sec-defs-AI} will provide definitions of other key terms used in the article,
including AI, data science, and knowledge.
The very general term ``AI'' will be used
to denote research and projects that involve the processing and analysis of knowledge and data, often with 
machine learning / data mining methods. This interpretation corresponds to the strong representation
of data science projects at least in the ``AI for Social Good'' literature, see Section \ref{sec-summary-of-findings}.
Specific references to data science and machine learning / data mining will be made when necessary.
               
Contributing to the Common Good 
is an ambitious and noble aim, and I am convinced that it inspires many researchers and practitioners to act in responsible ways. However, as I will argue in this paper, even with the best of intentions, certain characteristics of AI thinking and practice, coupled with the inherent need to act in politically charged environments, may impede ‘design for the Common Good’. 
To explain why, Section \ref{sec-questions} will detail four specific characteristics, summarized into four lead questions:
 the problem-solving and solutionism mindset of the engineer, the difficulties of integrating different stakeholders, the role of knowledge, and side effects and dynamics.  

Section \ref{sec-obvious} will validate the importance of the four lead questions via an exploratory survey of 99 contributions to
recent conferences on AI and Data Science ``for Social Good'' or ``for Good'', the notions that are currently most similar to
AI for the Common Good {\em and} that are sufficiently established to have formed conferences. 

Turning these challenges into a positive recommendation, the concluding Section \ref{sec-conclusions}  will draw on another characteristic of computer-science thinking and practice to make the impediments visible and attenuate them: ``attacks'' as a method for improving design. In analogy with penetration attacks, I will propose {\em ethics pen-testing} as a method for helping AI designs to better contribute to the Common Good. Further, I will argue why the arguments put forward here are characteristic of and relevant for AI and for the goal of enhancing the Common Good, but not restricted to the field or the goal.

\section{What is the Common Good?}
\label{sec-CG-and-terms}

\subsection{The Common Good as a goal for AI}

The ambition to be good for all (or at least many) people has become prominent throughout computer science in general and AI in particular. Some examples can be found in ethics codes:
\begin{itemize}
\item
ACM Code of Ethics and Professional Conduct \cite{ACM92}: ``1.1 Contribute to society and human well-being. This principle concerning the quality of life of all people affirms an obligation to protect fundamental human rights and to respect the diversity of all cultures.''
\item
Asilomar Principles \cite{Asilomar16}:
\begin{description}
\item[]``23) Common Good: Superintelligence should only be developed in the service of widely shared ethical ideals, and for the benefit of all humanity rather than one state or organization.''
\item[]``14) Shared Benefit: AI technologies should benefit and empower as many people as possible.''
\item[]``15) Shared Prosperity: The economic prosperity created by AI should be shared broadly, to benefit all of humanity.''
\end{description}
\item
Similar ideas are implicit in IEEE Ethically Aligned Design \cite[p. 5]{IEEE16}: the goal to ``develop successful autonomous intelligent systems that will benefit society'' and the second General Principle, to ``Prioritize the maximum benefit to humanity and the natural environment.'' 
\end{itemize}
The first thing to note in these different principles is how differently collectives are referred to. They range from 
‘not all of the benefits should accrue to giant internet companies’ (``rather than one state or organization'')
to literally ``all people'' or ``all humanity''. ``As many people as possible'' lies between these extremes, but is underspecified when one does not know what constitutes the possible. 
Further underspecified terms are the ``broadly shared prosperity'' and the ``widely shared ethical ideals'' (see Section \ref{sec-terminology-scope} for possible referents).
The wordings also leave room for different distributions of the benefits, and they make no statements about how to negotiate multiple and possibly conflicting ideals, values, and notions of what is good. 
Many of these questions have been and are being debated in the wider literature on the Common Good,
which is the subject of the following section.

\subsection{Some questions regarding the Common Good, inspired by the notion from political philosophy}
\label{sec-2-2}

The Common Good has been discussed widely and controversially by many authors in political philosophy, and it is impossible to survey this literature in the scope of this article.%
\footnote{The term is also used in different disciplines. For example, in economics, something is a common good if  no-one can be excluded from ``consuming the good'', regardless of whether it is a benefit or not. This concept relates to issues of access, distribution and scarcity of resources. The discussion of these issues would go beyond the scope of this article.} 
Instead, I will very briefly present some issues that raise relevant questions for the interpretation of the concepts proposed in AI.
 
The Common Good has been defined as ``that which benefits society as a whole'' \cite{LeeND}. But how are these elements (the ``that'', ``benefit'', ``society'') defined? 

Hussain \cite{Hussain18} gives more details about the {\em that}: ``the common good is […] part of an encompassing model for practical reasoning among the members of a political community. […] The relevant [interests and facilities that serve these interests] together constitute the common good and serve as a shared standpoint for political deliberation. […] The relevant facilities may be part of the natural environment (e.g., the atmosphere, a freshwater aquifer, etc.) or human artifact (e.g., hospitals, schools, etc.). But the most important facilities […] are social institutions and practices.’’ One example of such institutions and practices is a scheme of private property. Fundamental rights / human {\em rights} (``basic rights and freedoms’’) are parts of the Common Good \cite{Hussain18}. Finally, I will use {\em values} interchangeably with ``interests’’ for the purposes of the present article.

The notion of {\em benefit} also invites different readings: is it an individual’s or a group’s utility in a welfare consequentialist sense, and/or is it based on values beyond this? (Hussain \cite{Hussain18} favors the latter reading, but also reports on alternative conceptualizations of the Common Good.) Finally, the questions of what the boundaries of the relevant {\em society} (or:
political community and its members) are, and of whether to take a welfare consequentialist or other standpoint, and whether and how to account for collective above individual interests \cite{Blum12}, tend to receive less attention than others in ``for Good’’ initiatives, and will therefore not be considered further here.

But even the definitional elements of facilities, interests, and practical reasoning raise further questions. The following is a selection that 
contributed to the choice of lead questions proposed below.

A first questions is: Who defines the Common Good (or the interests and facilities) and how?
Political philosophy distinguishes between {\em substantive} and {\em proceduralist} conceptions of the Common Good. Substantive conceptions specify what factors, goods, values, etc. are beneficial and shared. Proceduralist conceptions instead focus on what procedures are adequate to collectively negotiate and define what is beneficial. 

The expression ``substantive value’’ is intended to denote the unassailable status of the value as something that can stand on its own and requires no justification. Yet that status is logically dependent on the attribution of the speaker, who categorizes the value as such. Any such self-supporting value is easily challenged by denying the attribution. Substantive values and their attribution have come under specific political and philosophical attacks after the atrocities of 20th century authoritarian regimes, who all professed to act in the interest of some common good, an ``attempt to make heaven on earth’’ that ``invariably produces hell.'' \cite{Popper}

Proceduralist notions of the Common Good rely on democratic structures and deliberation; it need  not be known a priori which facilities and interests will be agreed upon through these processes, see \cite{Jaede17}. Even if the focus of proceduralist notions is on process, this does not mean that there are no substantive elements, e.g. \cite{Blum12}. The need for substantive elements can arise from what Popper called the tolerance paradox (if a society is tolerant without limit, this tolerance can be abused or even destroyed by the intolerant). Countermeasures include constraints on the forms the deliberation can take (e.g., that citizens recognize each other as equal and use only reasons that can be accepted by all others \cite{Cohen}) and legal constructs that enable and require a country’s political bodies to protect the political order against those who want to abolish them, such as constitutional clauses that cannot be abolished even by a majority (``militant democracy’’, cf. \cite{doi:10.1146/annurev-lawsocsci-102612-134020}).

Another distinction is that between {\em communal} and {\em distributive} conceptions of the Common Good.  A communal conception takes the Common Good interests to be interests that citizens have as citizens, whereas a distributive conception is based on the acknowledgement that citizens belong to various groups with distinct interests, that these interests compete for the facilities and resources and may pose different demands, and decisions and allocations need to be made according to some distributive principle \cite{Hussain18}. 

\subsubsection{From questions about the Common Good to questions about AI for the Common Good}

The philosophical considerations about the Common Good that have been summarized very briefly in the previous section
served as starting points for the questions proposed in Section \ref{sec-questions}. Here, I will give an overview of
the link between the considerations and questions. 

The considerations above indicate that purely substantive accounts of the Common Good are problematic, that procedures are important,
and that groups
with different interests and demands may have different notions of the Common Good. These groups correspond to what computer science calls stakeholders.
These considerations were one inspiration for the first two lead questions: What is the problem, and who defines it (see Sections \ref{sec-Q1}
and \ref{sec-Q2})? A second inspiration was that these same two questions have proved constructive in interdisciplinary collaboration around a specific Common Good interest and facility: ``privacy’’
\cite{DeWolfEtAl16}. 

Note that the definitional duality of ``Good’’ and ``problem’’ introduced in the previous paragraph is frequent in AI: some value or aspect of the Common Good is missing, deficient, or
under attack, and this constitutes a problem. The problem then prompts a search for a technological contribution to solving or at least addressing this problem. 
The focus on a (usually technological) solution is the reason to ask a modified version of the first lead question: What is a problem in the first place (see Section \ref{sec-Q1dash})?

``AI for the Common Good’’ is understood here (and in the surveyed literature) in an engineering sense. 
Thus, AI methods, technology, and their deployment cannot be an interest, but a facility (or part of it) that serves an interest. This raises the question: what kind of facility is or should this be? I will argue that today, this is mostly some form of knowledge that is then fed into further decision processes. (Another candidate are autonomous systems, which would require a different analysis.) This inspired the third lead question concerning what the role of knowledge is (see Section \ref{sec-Q3}). 

The fourth lead question asks about important side effects and dynamics (see Section \ref{sec-Q4}). This can be related to the Common Good literature in that this literature also investigates what is likely to happen under different structures of people acting, deliberating, deciding, and collaborating. 

\subsection{AI and Data Science for (Social) Good}
\label{sec-AI4SG}
\label{sec-DSSG}

At this point in the argument, one would normally investigate how concrete projects or initiatives (rather than abstract ethics codes) define the Common Good. However, 
the call to develop and deploy AI for the Common Good has, to the best of my knowledge, not yet led to research programs or publications under that name.
At the time of writing of this article (August 2018), a Google search returned three texts. The phrasing ``AI for Common Good'' has been used in the titles of two recent white papers,
one prepared for attendees of the
2018 World Economic Forum Annual Meeting \cite{WEF18}, and one by North Highland Consulting \cite{NHC18}.
Both focus on highlighting threats posed by Artificial Intelligence; what the Common Good is and how to use AI towards it is not explicated.
In addition, an entry in the {\em Communications of the ACM}'s news, titled ``AI for the Common Good'' \cite{Mols17},
reports on the AI for Good Global Summit, whose goal definition is given below.

However, related ideas have a longer tradition and have led to several conferences and research programs.
To outline the field, I have selected four that I believe are most influential and representative
of ``AI for (some version of) Good''. The selection
was based on the duration of the initiative (at least two editions) and/or the backing by an important
professional association (AAAI) or an important international actor (the UN).
Since the initiatives present their definitions on their Web pages rather than in scientific publications, 
some interpretation is needed and will be supplied in the following paragraphs.

Initiatives around AI and Data Science will be presented, for several reasons. First,
``for Social Good'' originated as an initiative from data science, second, data science is one key
area of, or related to, current AI (for details, see the definitions in Section \ref{sec-defs-AI}),
such that, third, many contributions to conferences on AI for (Social) Good are or contain data science. 

The {\em Data Science for Social Good} (DSSG) initiative 
has organized, since 2013,
an annual ``summer program for aspiring data scientists to work on data mining, machine learning, big data, and data science projects with social impact. 
Working closely with governments and nonprofits, fellows take on real-world problems in education, health, energy, transportation, and more.''%
\footnote{\url{https://dssg.uchicago.edu/}. This description has been used since the first version of the site recorded in the Web Archive at 
\url{https://web.archive.org/web/20140318063609/http://www.dssg.uchicago.edu:80/}}
The first part of this definition is strictly speaking not very specific, since many uses of AI have social impact,
including large social networks, search engines, and (inter)national surveillance programs. 
This shifts the definitional core of ``Social Good'' to the domains of projects (such as education, health, energy and transportation)
and to the actors (governments and nonprofits) who are likely to be the ones to define project goals and/or who control and provide the data. 

A definition via {\em domains} and {\em actors} is also used by the Data Science for Social Good (SoGood)  workshop series that has so far seen three consecutive editions at ECML PKDD,
a major conference on machine learning, data mining, and data science:
 ``how Data Science can and does contribute to social good in its widest sense, including areas such as: 
    Public safety and disaster relief,
    Access to food, water, and utilities,
    Efficiency and sustainability,
    Government transparency,
    Data journalism,
    Economic development,
    Education,
    Social services,
    Healthcare.
We are interested both in non-profit projects and in projects that, that while not defined as non-profit, still have Social Good as their main focus, and so have managed to build a sustainable business model.''%
\footnote{from the first edition at \url{https://sites.google.com/site/ecmlpkdd2016sogood/}}

A focus on DSSG projects' {\em problem-solving} is suggested by \cite{TanweerEtAl17}: DSSG consists of ``attempts to solve complex social problems through the use of
increasingly available, increasingly combinable, and increasingly computable digital data''.

With a method scope of AI in general (rather than DS in particular), 
the Association for the Advancement of Artificial Intelligence held a spring symposium on ``AI for the Social Good'' in 2017.%
\footnote{\url{https://www.aaai.org/Library/Symposia/Spring/ss17-01.php}}
The AAAI Spring Symposia center on emerging topics in AI; hence, this is an indication of the endorsement of the field,
by a major professional association. ``AI for the Social Good'' is 
defined as AI ``addressing societal challenges, which have not yet received significant attention by the AI community or by the constellation of AI sub-communities, [the use of] AI methods to tackle unsolved societal challenges in a measurable manner.''%
\footnote{\url{https://aaai.org/Symposia/Spring/sss17symposia.php}}
Another venue defines the field by declaring ``almost any real-world problem, which is important for society’s benefit, and could potentially be solved using AI techniques, [to be] within the ambit of this symposium.''%
\footnote{\url{http://scf.usc.edu/~amulyaya/AISOC17/}} 
This definition reiterates the idea of ``benefit for society'', see Section \ref{sec-2-2}, and the focus on problem-solving.

With a method scope of ``Good'' in general (rather than ``Social Good'' in particular),
the ITU, the UN agency responsible for issues that concern information and communication technologies,
leads the ``AI for Global Good Summit''.
The Summit has so far been organized twice (2017 and 2018). The goal is described as 
``AI innovation [being] central to the achievement of the United Nations' Sustainable Development Goals (SDGs) by capitalizing on the unprecedented quantities of data now being generated on sentiment behavior, human health, commerce, communications, migration and more'', including goals such as ``no poverty'', ``zero hunger'', and ``good health and well-being''.%
\footnote{\url{https://www.itu.int/go/AIforGood2018}, \url{https://sustainabledevelopment.un.org/?menu=1300}}
These and most of the 14 other SDG goals have a {\em substantive} focus.

More specific societal goals, for example fairness (non-discrimination), are pursued by research communities such as Fairness, Accountability and Transparency in Machine Learning and beyond.%
\footnote{\url{https://www.fatml.org/}, \url{http://fatconference.org/}}
Another example is the protection of privacy as the goal of various research communities
including (in DS) privacy-preserving data mining and data publishing.
In addition, funding programs with similar goals exist. I have been part of two ``projects with a primary societal finality'' funded
by the Flemish Science Council FWO. In these projects, multidisciplinary consortia (of which AI was only one partner) investigated
privacy in online social networks and diversity in media, respectively. The methodological and ethical debates in these projects have been
an important source of inspiration for the current article.
Since these goals are quite distinct from (or at least much more specific than) the Common Good, these specific notions of the Good will not be investigated further here.

In sum, ``the Common Good'' is referred to as a goal for AI in current publications, but not defined.
Concrete current initiatives refer to the ``Social Good'' or simply ``the good'',
circumscribing it via problem domains and the identity of the non-academic project partners (nonprofits or governments),
or via substantive goals agreed upon at UN level. 
It appears that, as a common denominator, the intended beneficiaries of AI for Good, for Social Good, 
etc. will generally not be the ones who directly pay for the development or use of this AI.
 Thus, unlike for example in commercial application areas, no market price can serve as indicator of value.
``The Social Good'' is then the indicator of such value. 

Healthcare is an interesting example: while it can be a profit-oriented business model,
the focus in the domain of AI for (Social) Good appears to lie on the provision of healthcare to broader sections
of society (see ``for all'' above).
The historical experience suggests that such a provision requires some kind of national insurance financing scheme based on solidarity rather than payment-for-service.
Thus, is any contribution of AI to better healthcare methods already ``AI for Good'', or is more needed?
It is likely that such questions will be asked in the further development of the field.

\section{Terminology: AI, Data Science, knowledge, and ethics-in-AI}
\label{sec-defs-AI}
\label{sec-defs}
\label{sec-terminology-scope}

In this section, further key terms used in this article will be defined.

{\em Artificial Intelligence} (AI) is ``the ability of a digital computer or computer-controlled robot to perform tasks commonly associated with intelligent beings.
The term is frequently applied to the project of developing systems endowed with the intellectual processes characteristic of humans, such as the ability to reason, 
discover meaning, generalize, or learn from past experience'' \cite{CopelandND}.
In line with the ACM Computer-Science subjects classification, I regard AI as a subfield of computer science,
and as such a field at the intersection of science and engineering.
{\em Machine Learning} (involved in particular in the last three characteristics of the preceding list) 
is a field of AI, as encoded for example in the ACM Computer-Science subjects classification in its most recent (2012)
version.%
\footnote{\url{https://www.acm.org/publications/class-2012}}

{\em Data Science} (DS) is understood here as a subfield of AI, or more specifically as a field that (a) is situated in many universities within
AI groups and (b) draws heavily on methods developed or used in machine learning and data mining. (The machine learning aspect corresponds to 
the focus on learning models of data, and the data mining aspect to the focus on entire knowledge-discovery workflows.)
More generally, data science has been defined as ``the science (or study) of data'' and 
``a new interdisciplinary field that synthesizes and builds on statistics, informatics, computing, communication, management, and sociology 
to study data and its environments (including domains and other contextual aspects, such as organizational and social aspects) 
in order to transform data to insights and decisions by following a data-to-knowledge-to-wisdom thinking and methodology''
\cite{Cao17}. Conway \cite{ConwayND}, in a frequently cited online source, 
described data science by means of a Venn diagram: with regard to machine learning, data science
is situated in the intersection of machine learning and substantive expertise.

{\em Knowledge} is understood in two ways. On the one hand, I will refer to a notion of 
knowledge as used in psychology: ``a structured collection of information that can be acquired through learning, perception or reasoning''
\cite{pbook}, understood to be held by a human agent (mental representation). It is the knowledge about something in the world,
the expertise \cite{TanweerEtAl17} that generally draws on many sources and fields, such as different academic disciplines.
Such human knowledge can be both {\em input}
to a research or development activity, and its eventual {\em result}. 

On the other hand, I will refer to knowledge as the more immediate {\em output} of an AI or DS activity. 
Russel and Norvig \cite[p. 16]{RN} implicitly define knowledge as a structured collection of ``information [...] put into a form that
a computer can reason with''. The field of knowledge discovery from databases and the related fields of data mining
and data science focus on knowledge in the sense of ``novel, valid, potentially useful,
and ultimately understandable patterns
in data'' \cite{Fayyad96} -- where the intended recipient, who can use and understand these patterns as structured representations,
is often but not necessarily human. In all these meanings, knowledge is structured information, useful and/or understandable to a person or machine. 

In this article, the very general term ``AI'' will be used
to denote research and projects that involve the processing and analysis of knowledge and data, often with 
machine learning / data mining methods. This interpretation corresponds to the strong representation
of data science projects at least in the ``AI for Social Good'' literature, see Section \ref{sec-summary-of-findings}.
Specific references to data science and machine learning / data mining will be made when necessary.

A final note concerns the question: {\em Whose ethics}?

Robotics as such is not in the focus of the present article, but robots are relevant to the focus on AI and ethics.
Not all artificial intelligence is incorporated into robotics, and not all robots are artificially intelligent.  
However, the intersection is large and relevant, and I regard such AI robots as typical representatives of what the Asilomar Principles
call ``highly autonomous AI systems''. Regarding these,
three types of ethics are relevant.

The first type is the professional ethics of the researcher or practitioner (often referred to as computer ethics). This is guided by principles such as the Asilomar Principle
11) ``Human Values: AI systems should be designed and operated so as to be compatible with ideals of human dignity, rights, freedoms, and cultural diversity.''
Arguably, these ideals are widely shared, codified for example in the Universal Declaration of Human Rights.%
\footnote{Cultural diversity was not mentioned explicitly in the 1948 UDHR, but is implicit in its Article 27 when 
interpreted together with the 2001 UNESCO Declaration on Cultural Diversity, see \url{http://portal.unesco.org/en/ev.php-URL_ID=13179&URL_DO=DO_TOPIC&URL_SECTION=201.html}.}
Thus, one may interpret the ``widely shared ethical ideals'' of Principle 23) as consisting of these four, and possibly also others.
(According to Principle 23), ``[s]uperintelligence should only be developed in the service of widely shared ethical ideals 
and for the benefit of all humanity rather than one state or organization.'')

The second type is the professional ethics of a researcher or practitioner who develops robots. Following 
Veruggio \cite{Veruggio10}, I refer to this as roboethics: 
``Roboethics is not the ethics of robots nor any artificial ethics, but it is the human ethics of the robots' designers, manufacturers, and users.''
Asilomar Principle 10) constrains these design, manufacturing and use activities by positing 
``Value Alignment: Highly autonomous AI systems should be designed so that their goals and behaviors can be assured to align with human values throughout their operation.''

Note that this formulation focuses on the goals and behaviors of a robot and asks that these be aligned with human values (presumably those
of Principle 11) listed above), and that it does not make a commitment as to whether these goals and behaviors stem from the humans
designing, manufacturing, or using the robot, or from the robot itself. Thus, the question of whether machine ethics 
\cite{Asaro06, Malle16, WallachAllen08}
as a third type of ethics, 
the ethics of a robot (or in fact any AI system), exists and if so, what its properties are, is left open. In line with this, the remainder of this article will focus on computer ethics / roboethics as human ethics 
in the sense described above.

\section{How to create ``AI for the Common Good'': Four lead questions}
\label{sec-questions}

In this section, I will analyze four specific characteristics of AI thinking and practice that challenge and may impede design for the Common Good:
 the problem-solving and solutionism mindset of the engineer, the integration of stakeholders, the role of knowledge, and side effects and dynamics.

The questions will be illustrated by references to a running example from the domain of ``AI for Social Good''. 
The example itself is, intentionally, not a real example in the sense of being the contents of a specific AI paper, 
report or otherwise -- because the point of the present article is not to denigrate the merits of any particular project. 
Instead, the example is a fictitious synopsis of uses of AI/DS in various contexts, with these uses all focusing on the same issue.
The example centers on drugs, considered by some to be ``public enemy number one''%
\footnote{Richard Nixon’s statement about drug abuse in a famous press conference 1971, see transcript at \url{http://www.presidency.ucsb.edu/ws/?pid=3047}}%
, that is, the ultimate ``Common Bad'', whose absence would surely enhance the Common Good. 
While traditionally, such statements targeted illegal drugs, the recent US opioid crisis, which was declared a Nationwide Public Health Emergency%
\footnote{Donald Trump's declaration of 2017, repeated in 2018, see \url{https://www.whitehouse.gov/briefings-statements/president-donald-j-trump-combatting-opioid-crisis/}.}%
, has highlighted how a similar problem can originate from a substance that may be legally prescribed or 
illegally peddled. The opioid crisis also illustrates how public health and criminal justice issues continue to interact:
Within three paragraphs of one political speech, US President Trump lauded an initiative that caused people 
to turn 
in more than 900,000 pounds of unused or expired prescription drugs, and the arrest of criminal aliens with 76,000 charges and convictions for dangerous drug crimes.%
\footnote{see transcript at \url{https://www.whitehouse.gov/briefings-statements/remarks-president-trump-combatting-opioid-crisis/}}

The rhetoric around the opioid crisis has brought one question back into sharp focus: There is a problem, but what exactly is wrong? 
This leads to the first lead question Q1.

\subsection{Q1: ``What is the problem?''}
\label{sec-Q1}
Consider the example: What is ``the drug problem``? Here is a non-exhaustive list of candidates:
\begin{enumerate}
\item People use drugs. 
\item People sell drugs.
\item Certain people (e.g. the poor, black people, ...) use drugs.
\item Drug users commit crimes.
\item Drug users become homeless, ill, ...
\item Drug users die (earlier than they would have without drugs).
\item There aren't enough drugs available. 
\end{enumerate}
These alternative definitions are designed to represent, in a simplified way, the views of different people who are affected by drug usage and its consequences.%
\footnote{Of course, it is a simplification to equate such framings with actors; different interests, values, norms and
positionings further complicate the picture. Also, only digitizable formulations of the problem were included,
because others are outside the scope of an AI approach. Number 7 illustrates the contextual nature of problem definition:
This may be an addict's view when on withdrawal, even if the same person would under other circumstance worry more about the 
negative effects on health and life prospects.}
 If, therefore, a system to ``solve'' the drug problem, by AI or otherwise, is created, the question of who defines the problem counts.

\subsection{Q2: Who defines the problem?}
\label{sec-Q2}

A necessary condition for designing systems that further the Common Good is to hear the voices of multiple and diverse people who will be affected by the system. The integration of multiple stakeholders 
in requirements engineering
(for an overview, see \cite{ZwickerEtAl15})
 are therefore increasingly required by ethics codes as well as laws. Codes such as the ACM Code of Conduct
\cite{ACM92}%
\footnote{``3.4 Ensure that users and those who will be affected by a system have their needs clearly articulated during the assessment and design of requirements; later the system must be validated to meet requirements.''}%
, the AOIR  Recommendations for Ethical Decision-Making and Internet Research \cite{AOIR12}%
\footnote{cf. for example the sections on ``Key guiding principles'' and on ``Internet
specific ethical questions''}%
, and the IEEE Ethically Aligned Design guidelines \cite{IEEE16}%
\footnote{Section 2, Embedding Values Into Autonomous Intelligent Systems} 
explicitly call for this.
A special form of multi-stakeholder requirements elicitation has recently even attained the status of a legal obligation:
the European Union's new data protection law, the General Data Protection Regulation (GDPR), requires 
a
data protection impact assessment before personal data are collected and processed.
(While the current article focuses on ethics codes, I will make some references to the GDPR as an example of a current and wide-ranging
attempt to codify rules for technology, including AI, to protect individuals' rights and freedoms.)

As part of such processes, ``[i]t is understood that there will be clashes of values and norms when identifying, implementing, and evaluating these systems (a state often referred to as `moral overload')''  \cite[p. 23]{IEEE16}, so conflict resolution methods and processes are required. 
The study of {\em democratic} methods for gathering and negotiating requirements is a subfield of requirements engineering \cite{Goguen94}.
However, an ongoing challenge remains: how to best support
democratic deliberation and conceptions of distributive justice with software and/or software engineering methods.  

Conflict identification and resolution become more difficult when stakeholders are differentially able to cause a clash in the first place, because they are embedded in socio-technical systems differently and differ in their abilities to perceive and voice their values and norms. As an example, consider imprisoned drug users as one of the relevant groups in 
\cite{BirdEtAl16}. Further problems arise when affected communities and individuals are outside the boundaries of the society deemed relevant in 
the respective notion of the Common Good. For example, inhabitants of countries such as Colombia, in which drugs are grown and who suffer from the local effects of drug cartels' power, 
have argued that they are outside the consideration of drug consumers in the West \cite{Amrani18}. 

In addition, it is questionable whether talking to different stakeholders is enough, because it may not affect the structural mold into which these different stakeholders' utterances will be put: the very notion of what a problem -- any problem -- is. This will be investigated next.

\subsection{Q1': ``What is a problem?'' (and thereby: What is the problem -- revisited)}
\label{sec-Q1dash}

While conceivably no AI
researcher or practitioner would be as preposterous as claiming to ``solve the drug problem'', AI approaches do focus on a version of the problem (usually implicitly specified to be smaller).
This is evidenced by the above-mentioned reference to the use of AI on ``[a]lmost any real-world problem, which is important for society's benefit, and could potentially be solved using AI techniques''. This is germane to the discipline and the ``problem-solving mindset'' of the engineer. 

\subsubsection{Problem-solving}

In everyday language, a problem is ``a matter or situation regarded as unwelcome or harmful and needing to be dealt with and overcome''.%
\footnote{All definitions of ``problem'' from \url{https://en.oxforddictionaries.com/definition/problem}} 
In some cases, ``dealing with and overcoming'' may be relatively straightforward. Using another drugs example: if the problem is
a person's breathing being suspended due to an overdose, this problem can be dealt with and overcome by the proper administration of Naloxone.
However, most real-world problems, including many around drug overdosing, are more complex. 
For example, subjecting a heroin addict to a methadone program can ``deal with'' the heroin addiction, but it does not necessarily ``overcome'' it.
In addition, many real-world problems are open-ended. For example, there is probably no ``overcoming'' the fact that in any society, many people overuse or misuse legal or illegal drugs --
but this does not alleviate societies from the responsibility to ``deal with'' drug usage.

On the other end of the scale, there are chess problems: ``an arrangement of pieces in which the solver has to achieve a specified result'' or mathematics/physics problems: ``an inquiry starting from given conditions to investigate or demonstrate a fact, result, or law.''

Engineering problems may, but do not have to, start from the everyday notion --- but the engineering approach rests on transforming whatever the starting point is into a well-defined designation: moving from a stateA (undesirable situation) to a stateB (desirable situation) \cite{Kovacevic17}. Only once the ``problem'' has this well-defined shape, can the engineer begin to ``solve'' it. This is what makes engineering precise, graspable and powerful. 

However, most social problems are not chess problems, and they do not exist in context-free structures.
Therefore, 
en route to the definition of an engineering problem, one must usually make some assumptions that are hard or even impossible to formalize. 
Ex post, the ambivalence that existed in the beginning tends to be cognitively minimized and the result is taken to be {\em the} truth (even if it did result from decisions that might just as well have been made otherwise), cf. \cite{Naur82}. 
The result is, partly, that the definition of the problem often arose from the opinions of only one or only a few stakeholders. Multi-stakeholder methods and participatory software design are approaches for addressing this issue. However, regardless of which and how many stakeholders have been consulted, formalization remains a necessary step.
The risk of conceiving of social problems in terms of engineering problems is to blind oneself to the vagaries of the formalization step, 
and to fail to consider alternatives to the chosen formalization.

\subsubsection{Problems and ``solutions''}
 
Likewise, what do we consider as ``solutions''? This will depend on the context in which problem-solving takes place. Consider several standard problem-solution pairs, with solution approaches from law, law enforcement, and public health, in different countries. The current discussion will focus on illegal drugs. In general, problems 1-7 above exist for legal drugs too, but are addressed differently.

In the War on Drugs that began in the Philippines in 2016, attention has focused on the first and second of the problem versions from Section
\ref{sec-Q1} above (henceforth, \#1 and \#2). The ``solution'' proposed in the election campaign of President Duterte as well as enacted in a large number of cases was to stifle both supply and demand by killing drug dealers and drug users, cf. \cite{Villamor17}. Another ``solution'' approach consists of criminalization and incarceration laws and policies for drug dealing and use, even of small quantities, as in the US during recent decades. Some countries exempt the ownership and consumption of small quantities of illegal drugs from prosecution (i.e. prioritize \#2 in the problem definition). 
Various societal actors and authors (e.g. \cite{ACLU14,Gwynne13}) have argued that 
problem definition \#3 stands behind US laws that penalize the use and dealing of drugs traditionally associated with poor and black users (crack cocaine) far heavier than that of similar drugs traditionally more prevalent among affluent white users (powder cocaine). 
The identity of drug users in problem definition \#3 can even become associated with a \mbox{(re-)}framing both of ``the problem'' and ``the solution''. This point has been made after US President Trump, in October 2017, declared the opioid crisis (which affects many poor white people) a public health emergency rather than another type of drugs on which to wage war \cite{Mechanic17}.

\#4 can be a problem in at least two ways: \#4.1 when drug users commit crimes such as theft or prostitution to finance their addiction, or \#4.2 when intoxicated people become uninhibited and/or aggressive and then commit crimes such as assault and murder. \#4.2 is an often-voiced observation concerning the legal drug alcohol, and ``solutions'' contain penalties for driving when intoxicated. (Beyond that, for legal drugs there tends to be a separation between permitting the intoxication as an expression of personal freedoms, and the sanctioning of crimes if they occur.) \#4.1, on the other hand, is in many cases tightly linked with \#5, and programs such as substituting methadone for heroin in (a) legal, (b) insurance-covered, and (c) medically administered ways are
offered as partial ``solutions''. If \#6 is considered the main problem, the question for a solution may need to turn from the ubiquitous attempts to reduce consumption to accepting the fact that consumption and overdosing happen, and counteract the lethal effects of overdoses as they appear. 

These approaches from law, law enforcement and healthcare are not, or not in their entirety, AI-based. AI is used to address parts of the problems, as when predictive policing software recommends where to patrol for drugs \cite{LumIsaac16}. The AI models underlying such software need well-defined and accessible data and well-defined objective functions whose maximization constitutes a ``solution''. This can lead to police deciding to patrol where there have been many drug-related arrests in the past, rather than where there has been much drug usage, and it will commit countermeasures against such biased decisions to one formalization of ``fairness'', which may mean that other notions of fairness are violated \cite{Chouldechova17,KleinbergEtAl17}.
Such side-effects of AI ``solutions'' should be kept in mind. To keep the unavoidable restrictions more visible, references to ``problems'' and ``solutions'' should be replaced by terms that are more clearly technical and limited in scope, such as ``the task to be done''. In the example, the task could be to decide where to patrol. 
Once the task is clear, the question becomes what to do or communicate. 

\subsection{Q3: What is the role of knowledge?}
\label{sec-Q3}

Above, I have argued that the transformation of a social problem into a formal problem poses challenges when the goal is to contribute to the Common Good.
In this section, I will study two challenges related to this transformation, both of them related to knowledge.
The first concerns the knowledge that enters (or fails to enter) into the transformation into a formal problem.
The second concerns the reverse direction: what happens when 
the output or ``solution'' of this formal problem is a piece of generated knowledge?
And what side-effects arise when both input and output are (necessarily) entangled with the knowledge of the AI developer, funder, ...?
How may AI methods themselves affect such knowledge effects?
What consequences may these constellations have on the Common Good? 

AI and in particular DS are strongly linked to knowledge:
The goal of AI is often described in a procedure-oriented way, such as in the definition presented in Section \ref{sec-defs}: 
to ``[develop] systems endowed with the intellectual processes characteristic of humans, such as the ability to reason, 
discover meaning, generalize, or learn from past experience''. Yet, early on in the history of AI, it became clear that 
only focusing on algorithms (e.g. for reasoning) will not produce intelligence, but that strong knowledge bases are
also needed. Terminology and valuations of logics-based approaches change over time, and currently, the reliance on 
knowledge is often expressed as a reliance on (big) data instead. In other words,
AI systems and their outputs are considered useful if they work on large and rich data or knowledge,
and if their outputs present new information that humans derive further knowledge from and act upon.%
\footnote{The distinction between data, information, and knowledge is a classical discussion in AI, but since the focus
here is on the practical consequences of deployed AI technology rather than on its philosophical foundations,
the discussion is not relevant for present purposes.}

\subsubsection{The  power of knowledge: AI and framing effects}

The framing of a problem denotes the way it is described. 
The framing of a problem (and its associated ``solutions'') has a powerful effect on how people -- and thus the public as those who may benefit, or be harmed by an AI system -- perceive the world and act in it, e.g., \cite{TewksburyScheufele09}. Frames can themselves be objects of knowledge, that is, statements can be made about them and discussed. For example, different frames can be identified
and compared as alternatives.
Designers should be aware that their knowledge-based methods operate in an environment filled with politically and otherwise induced frames and that association-based methods tend to reinforce these frames.
 These frames come loaded with certain versions of the notion of the Common Good (and many more blind spots regarding it). To work in the interest of the Common Good, an AI researcher
or practitioner
 should be aware of this fact and make conscious (and transparent) choices about whether to sustain frames or expose them.

Framing interacts with AI for the Common Good on multiple levels. 
First, frames contribute -- or could contribute -- to problem definition.
Second, AI, by the way it processes knowledge, can serve to reinforce such frames. 
Third, frames operate not only on the level of problems and solutions, but also on the level of methods.
These effects will be considered in turn.

Frames emphasize certain aspects of reality, and they suppress or even block others. One consequence is that specific views -- including but not limited to the perception
of what the problem is by
specific stakeholder groups -- can remain suppressed, or that the knowledge that certain solutions do not work is blocked. 
            
An example is the trope of the ``War on Drugs''. Launched in 1969 by then-US president Richard Nixon, this may have been influenced by concerns over public health and the suffering induced by abuses of illegal drugs. However, an alternative view has co-existed with this, succinctly described by Nixon's then counsel and Assistant to the President for Domestic Affairs: ``The Nixon campaign in 1968, and the Nixon White House after that, had two enemies: the antiwar left and black people. [...] We knew we couldn't make it illegal to be either against the war or black, but by getting the public to associate the hippies with marijuana and blacks with heroin, and then criminalizing both heavily, we could disrupt those communities. We could arrest their leaders, raid their homes, break up their meetings, and vilify them night after night on the evening news. Did we know we were lying about the drugs? Of course we did.'' \cite{Baum16}%
\footnote{This has not gone undisputed, see \cite{Hanson16}.}%
. Baum's view changes the roles of pieces of knowledge: 
In the War on Drugs frame, the drug-taking is the problem, and the solution is (or at least involves) arresting people, raiding their homes, breaking up their meetings, and using the media to vilify them.
In the Baum frame, the existence or growing influence or expectedly growing influence of antiwar and black citizens is the problem, and an (at least intermediate) solution is to wage a War on Drugs.
Independently of whether one follows Baum's sinister interpretation of the political will of the Nixon campaign, there is now a more widespread acceptance of criminalization and incarceration not having ``solved'' anything at all \cite{Baum16}: (a) legal alternatives to criminalization exist, have been or are being tested in different countries, and have often led to measurable improvements in health and crime statistics, (b) legal drugs (in particular alcohol) and prescription drugs (such as opioid painkillers) affect far more people than illegal drugs and cause enormous human suffering and economic costs, and (c) the ``cure'' of criminalization imparts suffering and creates new problems, also for the Common Good. 

Frames tend to be reiterated and ``echoed'', and they can survive even in the face of clear scientific evidence that contradicts a frame.
Arguably, much of current drug policy rests on reiterated, ``echoed'' frames that have been constructed and perpetuated for a number of decades now. One example is the perpetuation of the focus on criminal justice even in the context of the National {\em Public Health} Emergency proclaimed by US President Trump, see above. Another example is the controversy around David Nutt's two studies in {\em The Lancet} arguing that alcohol and tobacco are far more dangerous than common illegal drugs. Among other things, Nutt was dismissed 
from the UK government Advisory Council on the Misuse of Drugs, see the summary and links in \cite{WP-Nutt}.
 
Mass media have long been known to be influential in relaying, reinforcing and maintaining frames \cite{Ewen96}, creating ``echo chambers''. 
Social media have recently been described as intensifying the echo-chamber effects that media often have anyway, e.g. \cite{BakshyEtAl15,DelVicarioEtAl16,WilliamsEtAl15}.
This is where AI enters: recommender systems (the backbone of modern social media platforms' approach to addressing information overload) work on associations learned from past data and thereby tend to further strengthen these effects, e.g. \cite{NguyenEtAl14}. 

Framing operates not only at the level of problems and solutions, but also on the method level.
This can have wide-ranging effects on decisions for example in data-science related projects.
For example, structuring a decision-making process and tool by first identifying potential harms at a general level
and then weighing them against specific and contextual potential benefits ``would always, it seemed, come out in favour
of intervening and therefore in favor of the data sharing that would enable intervention'' \cite[pp. 4-5]{Taylor1}.
In an analysis of big data approaches to epidemics in high-income vs. low-/middle-income countries, 
the conception of populations as well-informed individuals versus as pathogen-carrying groups may imply that
``big data models built to facilitate individuals' well-being and autonomy instead would constitute perfect tools for mass control and
surveillance'' \cite[p. 30]{Taylor2}. 
Finally, the correlation-based nature of data mining itself has effects. Data science models used in the criminal
justice sector have been criticized widely for their effects of reproducing societal biases against minorities, cf. the
proceedings of the FAT(ML) conferences as a specific branch of AI for Social Good, see Section \ref{sec-AI4SG}.
As Barabas et al. \cite{Barabas18} point out, there are however risks for all, regardless of minority status,
when the underlying epistemic assumption is that persons ``have'' criminal tendencies innately correlated to their features 
and the intervention focusses on decisions about bail or incarceration, rather than on causal factors and the possibility
to change them via effective diagnosis and intervention of criminogenic needs. 

In sum, frames affect AI projects, and they can be reinforced by AI techniques. Framing may be unavoidable,
but it needs to be reflected and -- if appropriate -- counteracted.
Methods such as those designed to detect frames and bias, and to increase diversity in recommendation \cite{KunaverPozrl17}, can be components of
a more critical stance towards these mechanisms.

\subsubsection{Limits of imparting knowledge: does it work?} 

Given the strong effect of framing, someone who ignores frames and believes that they 
can immediately discern ``the facts'' in information they consume, or can immediately convey ``the facts'' in information they produce and communicate,
{\em underrates} knowledge. At the same time, many
presentations of AI tend to {\em overrate} knowledge in the sense that they suggest that the existence of knowledge can by itself solve problems. 

Such overrating can occur when knowledge is imparted for awareness raising. This is the goal of awareness tools in general (see \cite{BerendtEtAl14} for an overview specifically with regard to privacy awareness tools) and today is found in many quantified-self apps.
Basically, everybody (including drug users) knows that drugs are bad for health, family life, socio-economic status, etc. -- but this does not stop an addict from consuming the drug when the opportunity arises. In general, the limitations of ``just informing'' have been studied intensively in recent years by, for example, behavioral economists, and alternatives such as ``nudging'' have been investigated. These approaches rest on acknowledging that ``knowledge is not all'' and that decision-making is influenced by a wider range of factors than just classical rationality.  
One important group of factors are social influences. Many quantified-self apps and related applications, including
those around substance abuse or addiction problems%
\footnote{Examples are described at
\url{https://www.addiction.com/12575/six-sobriety-apps-you-should-know-about/} , \url{https://www.technologyreview.com/s/604085/treating-addiction-with-an-app/}, and \url{https://saferlockrx.com/top-apps-that-help-parents-prevent-teen-drug-abuse/}}%
, try not only or not at all to impart knowledge, but rather to help build and maintain social-support groups.

In their focus on using IT as communication technology, many of these apps are not AI-based. So where does AI come in? A very good example is Bird et al.'s
\cite{BirdEtAl16} use of data science methods informed by definition \#6 of ``the drug problem''. The authors used a data-science analysis to show the prevalence of lethal overdoses among addicts recently released from prison (at which time addicts are even more vulnerable than usual due to the enforced abstinence while in prison), and then instead of arguing for an awareness campaign ``to avoid drugs'' or ``avoid overdosing'', handed out emergency overdose kits and gave basic information  on how to administer the antidote. They showed, again with methods from data science, the effectiveness of their intervention: the number of deaths decreased significantly. 

All these ``solutions'' rely on certain definitions or framings of problem and solution, and all of them depend on various factors determining decision making. Data scientists and AI designers can draw on methods for supporting these factors and decisions studied in human-computer interaction \cite{JamesonEtAl14}, but as the Bird et al. example suggests, they should also be ready to think outside the box and embrace solutions in which they may only shine as diligent data analysts in the background, rather than as providers of smart knowledge-based tools in the foreground. 

But even when imparting knowledge works, is it always a good thing?

\subsubsection{Limits of imparting knowledge: is it good?}
AI and in particular DS often appear to operate on the assumption ``The more knowledge, the better``. 
This idea is applied to individuals as potential holders of knowledge, and it 
also impinges on the idea of the Common Good: ``The more knowledge society has, the better''.
But is this always the case?

At the individual level, there are certain well-known problems. First and as argued in the previous section, imparting knowledge may not work in the envisaged way. It may also manipulate people and negatively impact their autonomy, or hurt them in other ways \cite{DeWolfEtAl16}. It may place undue burden on them by making them responsible for tasks they lack the mental, financial, temporal, etc. capabilities for (``responsibilization'', \cite{Shamir08}). A growing number of legal and ethics guidelines recognize such limits. These include culturally-grounded restrictions on imparting knowledge \cite{RockwellBerendt17} as well as ``the right to not know'' in bioethics. Not knowing certain things is also recognized as a helper against unconscious biases, and it can therefore have economic advantages. The properties of such forms and conventions of not-knowing are investigated in the field of ignorance studies \cite{GrossMcGoey15}. 

If imparting knowledge is not necessarily beneficial, a parsimony principle can be useful: focus on the task at hand and get and use the knowledge needed for it, but not more. This principle is inspired by ethical, legal, and general intellectual principles. The Nuremberg Code, an early and highly influential code of research ethics, 
posits:  ``The experiment should aim at positive results for society that cannot be procured in some other way.''%
\footnote{cited after \cite{WP-NurembergCode}}
 The legal principle of proportionality (which pervades laws in general, and is particularly clearly adaptable to current purposes when a knowledge-based activity interferes with the fundamental right to data protection) says that the measure should be necessary to reach the goal. Similarly, in data protection principles and laws such as the GDPR, data minimization (collecting and using as little personal data as possible for the task at hand) is a guiding principle. Finally, arguably minimalism is considered a scientific virtue, expressed by general principles such as Occam's razor down to specific topics such as zero-knowledge proofs. 

\subsection{Q4: What are important side effects and dynamics?}
\label{sec-Q4}
Although computer science arose from cybernetics, the study of systems and feedback loops, much of today's computer science rests on surprisingly linear and short-term cause-effect relationships. This is probably due to that other basic principle of the natural and engineering sciences: divide and conquer, that is, split problems into parts and address these separately. However, side effects and dynamics of applications are becoming more visible. Again, these may negatively impact the overall effect of AI systems and thereby reduce, annihilate, or even reverse positive effects on the Common Good.

\begin{table*}[t]
\begin{tabular}{ l l l }
\hline
~ &
{\bf Engineering} &
{\bf Dealing with social problems} \\
~ &
{\bf problem solving approach} &
~ \\
\hline
Goal function &
well-defined & 
often involves unresolvable\\
~ &
~ & 
ethical dilemmas,\\
~ &
~ & 
continued re-negotiation \\
Method &
can be black box &
requires fairness, transparency, \\
~ &
~ &
accountability \\
Decomposition &
modular &
usually interdependent \\
Delegation &
can be fully delegated &
(at least some) participation required \\
Ethics codes and assessments &
canonical, &
starting points\footnote{``We must use critical thought to distinguish what is ethical from what is lawful and to consider what it means to
be a professional. Therefore, we must continually question and re-question authority, whether it is the law or a
code of ethics, or else we may be doomed to serve the interests of those who crafted the code, not necessarily the
interests of those who need to embody the code or use it to guide their practice. Just because a principle is
codified does not make it ethical.'' \cite{PatelElkin15}}, guidelines\footnote{``Multiple judgments are possible, and ambiguity and uncertainty are part of the process. We advocate guidelines rather than a code of practice so that ethical research can remain flexible, be responsive to diverse contexts, and be adaptable to continually changing technologies.'' \cite{AOIR12}}
\\
~ &
often a burdensome afterthought &
~ \\
Solvability &
Problems can be solved. &
Some problems can only be addressed.\\
\hline
\end{tabular}
\caption{Solving engineering problems vs. dealing with social problems}
\label{table-1}
\end{table*}

I will illustrate these with some examples from our example domain.

First, by any version of ``the drug problem'' relating to people and their behavior, inevitably many personal data will be collected and processed. This raises data protection issues that could well outweigh any positive benefits.
As an example, consider social network mining, which is currently a popular method also for studying drug usage, see for example \cite{YakushevMityagin14}.
In the event of a data leak%
\footnote{Social network platforms and apps, and quantified-self apps (see above), are often not very secure. Data and system security, while key in the GDPR and also mentioned in some AI ethics codes but not in others, does not receive as much attention in AI as it probably should.}%
, such social network mining methods could be used to derive inferences about individuals' drug-related behaviors or propensities that may damage reputations and
affect lives. Alternatively, social-media users could receive targeted advertising based on their supposed propensities, and vulnerabilities could be exploited.
Importantly, the question is not so much whether the mining methods return true knowledge, or whether the validity of the targeted advertising can be demonstrated:
the attempt at manipulation itself may present the problem. This is a lesson learned from the history of the Cambridge Analytica case, in which ideas from an academic project
in which social media were mined to predict personality
\cite{Kosinski13} were later supposedly used for psychometrically micro-targeted election advertising \cite{CAsummaryFromGuardian18}.

Second, it is by now well-known that ``objective'' big data analyses are likely to reproduce the biases in the data they learn from (thus violating the right to non-discrimination) \cite{BarocasSelbst16}. This has been argued for a wide range of big-data applications \cite{ONeil16} and shown with simulations for example for drug patrols \cite{LumIsaac16}: A predictive-policing application learns from past data that arrests have occurred frequently in certain areas, and it proposes that police patrol these areas preferentially. This leads to more arrests in these areas, which in turn feeds the learning to 
 propose to patrol them, etc. In general, since the deployment of big-data analyses will itself create data that then become input to further data analyses, this can easily create vicious-cycle phenomena that have been observed by sociologists for long, dynamics that can perpetuate or even aggravate bias and discrimination \cite{BerendtPreibusch17}. 
A promising research direction for breaking such feedback loops, drawing on reinforcement learning, proposes a different strategy for patrolling that could help to also detect the
cases in the (initially) less likely areas \cite{EnsignEtAl18}. It will be interesting to see how such strategies can be put into practice, and what effects this will have.

Third, self-reinforcing feedback loops can occur not only at the level of data, but also regarding technology use. Problematic drug usage is a form of addictive behavior. But if someone decides to control their drug use through an app, this can contribute to addictive forms of Internet usage, and the question arises whether this substitution is sensible. Vice versa, learning to use an app ecosystem in a way that is conducive to one's well-being, could help overcome substance abuse. It is an open question which factors contribute to these dynamics playing out in vicious or virtuous feedback loops.

\subsection{How a Solutionism mindset may hinder the asking of these questions}

``Solutionism'' is a term coined by Morozov \cite{Morozov13}. One of its definitions is ``the belief that all difficulties have benign (usually technological) solutions''. It can be regarded as an outcome of the problem-solving mindset described above, but other issues described under Q1-Q4 also play a role.
In their article on lessons learned from successful DSSG projects, 
Tanweer and Fiore-Gartland stress the critical importance of expertise on context, project partner organizational culture, and multiple stakeholder perspectives.
They conclude that  
``exposing an inequity or proposing a solution to a social problem doesn't
necessarily mean that social good will follow. If we ignore that warning, we are in danger of lapsing into
technological solutionism (Morozov, 2013), where we propose data-informed solutions that have little
chance of actually making a difference because they are contextually misconstrued, organizationally
untenable, or socially unacceptable.''  \cite[p. 3]{TanweerEtAl17}

Table \ref{table-1} juxtaposes relevant aspects of engineering and social-science mindsets. Solutionism, in this table, is the assumption that a social problem (which usually resides on the right-hand side) is a problem on the left-hand side, coupled with the associated treatment of this problem. 
The point is not to declare the problem solving approach as useless for AI striving for the Common Good -- on the contrary, its clarity and explicitness can often 
prove highly beneficial for method and system development.
Also, AI researchers and practitioners increasingly work in interdisciplinary teams, and sometimes also draw on skilled decision analysts who are much more alert to the complexities of decision making. However, the temptation to consider problems ``solved'' by a technological ``solution'' remains strong, and it can stand in the way of seeing and addressing the wider social issues.

\section{Is the need to ask these questions not obvious? An exploration of current publications}
\label{sec-obvious}

Some readers may concur that Q1-Q4 are important, but ask:
Is it not the case that all computer scientists and AI researchers and practitioners know how important problem formulation is? Is it not the case that they are all aware of the central role of stakeholders, and different stakeholders, in software requirements engineering? Is it not the case that AI researchers and practitioners know about limitations of knowledge, and that computer scientists, coming from a field that has its roots in cybernetics, are aware of systems and their dynamics? 

To get a first indication as to how AI in the interest of the Common Good deals with these aspects, I turned to four major venues for AI / DS for (Social) Good. The reasons for this choice are the same as for their choice as providing definitions, explained in Section \ref{sec-DSSG}. 
In these conferences, a large number of impressive methods and projects were described.
The following analysis in no way intends to downplay these approaches' positive contributions,
and caveats with regards to the study's results (which may derive from the conferences' goal
being ``Social Good'' rather than ``the Common Good'') will be described in the discussion in Section \ref{sec-5.3}.

\subsection{Surveyed materials}

I consulted all hyperlinked contributions (articles, extended abstracts, and presentations) that are made available on the websites of the four venue’s most current editions or in their published proceedings. This procedure gave rise to 
24 extended abstracts or articles from the Data Science for Social Good Conference 2017%
\footnote{\url{https://dssg.uchicago.edu/data-science-for-social-good-conference-2017/agenda/}}%
, 4 articles from SoGood 2017 \cite{SoGood17}, 15 articles from the 2017 AAAI Spring Symposium 
on AI for Social Good \cite{AAAI-SS-AISG}, and 56 presentation slide sets, 
extended abstracts or articles from the 2nd AI for Good Global Summit%
\footnote{\url{https://www.itu.int/en/ITU-T/AI/2018/Pages/programme.aspx}}%
. 
The selection contained all hyperlinked contributions to the second, third and fourth of these venues, since they all presented projects or methods (introductory greetings and other organizational materials were not considered further). The first venue, the Data Science for Social Good Conference, had multiple tracks, of which three appeared pertinent to the present questions and were therefore analyzed: DSSG {\em Fellowship Project Talks}, 
{\em Short Talks: Research Challenges in Doing Data Science for Social Good}
and {\em Short Talks: Collaboration Models for Social Good}. All contributions were read and assessed with regard to content and whether they contained explicit information relating to 
the four lead questions above.

\subsection{Summary of the findings}
\label{sec-summary-of-findings}

\subsubsection{Content} The contributions covered a wide range of issues, with no specific issue covered by more than one article. The motivation was usually framed as a problem to be solved; in some cases, a general social-good goal was named or could easily be inferred from the introduction and the specific computational goal. The issues were, in the large majority of cases, of a {\em substantive} nature. To the best of my knowledge, no generally agreed-upon ontology of Social Good objectives exists; I have therefore performed a rough classification by the SDG goals that were chosen as the guiding principle of the AI for Good Global Summit, and will report only the most frequent ones. 
I will in some cases distinguish between contributions to the first three venues and those to the fourth, for two reasons:
the Summit contributions were mostly slidesets that presented little method detail (such that counts of some methodological questions not being covered may be misleading);
and the Summit was structured into four thematic tracks (such that counts of topics in this conference follow from this structure, which is not the case in the other conferences).  

Sustainable Development Goal (SDG) \#3, Good Health and Well-being, was the most frequent: Eight out of the 43 contributions to the first 3 venues covered topics related to this SDG.
The topic was also covered in eleven of the contributions to the fourth venue, in which {\em AI + Health: Artificial Intelligence -- a game changer for Universal Health Coverage?} was one of the four tracks into which the conference was structured. Five contributions in venues 1-3 could be linked to SDG \#11, Sustainable Cities and Communities, and two to transportation (which concerns both cities and SDG \#9, Industry, Innovation and Infrastructure). In addition, a second track in the AI for Global Good Summit with 13 contributions focused on {\em Smart Cities}. The AI for Global Good Summit dedicated a third track, with 6 contributions, on {\em AI and Satellite Imagery} and linked this method-centric topic specification for contributions to three further SDGs (No Poverty, Life on Land, and Zero Hunger), leading to a strong representation of these topics. 

Only a minority of the contributions focused on {\em procedural} issues. Of these, five were concerned with the scientific process as such,
such as crowdsourcing a health-related task to citizen scientists.
These contributions already start from the assumption that the goal of the respective scientific project is a legitimate goal for the social/common good (``the aim is not to promote user interaction but to collect useful data for their scientific goals’’ \cite{Bartumeus17}). The fourth track of the AI for Good Global Summit, {\em Trust in AI}, in its descriptions and contributions likewise considers the goodness of AI as a given and the need to build trust as a way to convince people of this. Four contributions made proposals for the processes of working towards the Social Good, ranging from DSSG projects via non-profits to UN agencies tasked with identifying SDGs in national development plans. 

Three contributions dealt explicitly with democratic processes (in which citizens deliberate about their visions on the Common Good): one presenting a case study platform to create a democratic city planning system \cite{Bria17}, one presenting a case study platform to help make city growth equitable by increasing transparency and accountability \cite{AuerbachEtAl17}, one proposing an agent-based architecture to predict the effects of policies \cite{DignumDignum17}, and one presenting a mathematical voting model \cite{Prasad17}. 

Another goal that could be linked to processes was to improve information access/diffusion and quality (five contributions): the proposed analysis methods for summarizing news and social media contents and identifying misinformation, can arguably help citizens make better-informed democratic choices. 

\subsubsection{Q1: Alternative goals?} The vast majority of contributions worked with one (sometimes vaguely described or even only implicit) social goal and one computational goal. Thus, there was in general no attempt at framing the social problem in different ways (see Q1 above), and no discussion of whether and how the selected social goal could translate into different computational goals. 
The difference between the two types of problems (Q1’) was not the topic of any paper.

\subsubsection{Q2: Different stakeholders, and a description of how their perspectives and needs were assessed?} The large majority of contributions took their problem definition from the non-academic project partner, which was usually also the data provider (for example, a city council, a health agency, or a transportation agency). 
In some cases, several partners were mentioned (e.g., a tourism agency and a transportation agency), but no conflicts of goals or problems were reported. 
Tanweer and Fiore-Gartland, in their report on best practices, mentions stakeholders repeatedly, differentiating between ``partner organizations’’ and ``affected communities’’: ``DSSG projects can be more effective when done with consideration for the structures and cultures of partner organizations. This knowledge is often tacit for those stakeholders […] but without it, DSSG teams run the risk of developing products and services that have little chance of being embraced by stakeholders […]  DSSG teams need to view social issues from multiple perspectives, realizing that different communities and interest groups have [different and] sometimes conflicting stakes in the way social problems are portrayed and addressed. Without understanding the complex political landscapes and contested histories within which social problems are enmeshed, they run the risk of alienating affected communities’’ \cite[p. 3]{TanweerEtAl17}.

Only one contribution described an explicit multi-stakeholder process that was used to formulate the computational/engineering problem \cite{Hsieh17}, and another one mentions that their project partners followed such a process, in which there are clearly visible different positions \cite{AuerbachEtAl17}. 
An agent-based architecture to reflect different positions and interests was proposed in
\cite{DignumDignum17}, but the question of how to elicit these positions and interests was left implicit in this paper. 
In two contributions, differences between stakeholders’ interests are identified, but only one position is then pursued in the method or tool \cite{Guerin17}, or the problem is delegated by proposing that the owner of the AI device chooses the position that the machine will follow \cite{Bendel17}. 
Two contributions dealing with questions of fairness take the existence of different viewpoints of what ``fair’’ means as the starting points of their formal models \cite{Baumgartner17,Gummadi17}. The contributions dealing with democratic processes, especially \cite{Bria17}, implicitly acknowledge different viewpoints, but do not provide any specifics.

\subsubsection{Q3: The role of knowledge} It is difficult to describe the breadth and depth 
of knowledge that was {\em brought to} the contributions, since that would require an in-depth understanding of all the
domains of all the papers, or at least a validated bibliometric method. Both of these are beyond the
possibilities of the current article. However, it can be observed that the setup of the venues strongly encourages the participation
by AI researchers and practitioners, if only because the venues are defined in a discipline-centric way (``AI for ...'', ``Data Science for ...'').
This limits the incentives for people from other fields to participate. 

Concerning the role of knowledge as an {\em output of} the contribution, the texts give clearer indications.
First,
data science methods were not only the subject of the ``Data Science for ...'' venue, but also of many contributions to the ``AI for ...''
venues (solely or, for example, in combination with computer vision in the AI and Satellite Imagery Track of the AI for Good Global Summit). 
This leads to a strong representation of knowledge-centric methods.
Second, nevertheless, 
eleven contributions are coupled with an explicitly identified and specific intervention, such as apps designed to detect health problems \cite{Weber17} or apps to incentivize people to cycle to work \cite{Santo17}. A further contribution mentions that the project partner intends to use the developed tool for a number of specified purposes \cite{Scott17}. Two contributions \cite{Liu17,TanweerEtAl17} describe collaboration processes and thereby go beyond knowledge. For several contributions in the AI for Good Global Summit, it was difficult to see from the slide-set presentation what roles knowledge and interventions played. Thus, the numbers given here are likely to be a lower bound on those contributions that went beyond knowledge.

\subsubsection{Q4: Dynamics} Auerbach et al. \cite{AuerbachEtAl17} interleave data analysis and policy intervention and build a tool to satisfy various information needs in this process. They include a discussion on how insights from their data analysis and possible actions based on these insights could interact in the future and what this would imply for applicants and their use of the financing instrument they study. In the studied sample of contributions, this was the only one that included an explicit consideration of possible dynamics. 

\subsection{Discussion}
\label{sec-5.3}

The results indicate that so far, the considerations presented in the current paper are not an integral part of current operational research practices in AI/DS for (Social) Good. On the other hand, the importance of Q1 and Q2 is stressed in many methodological papers in the analyzed publications sample.
A question akin to Q3 is also highlighted from within the DSSG community: the importance of broader, and experiential, knowledge.
Q4 is, at the moment, mostly reflected in the fairness/non-discrimination literature. 
On the other hand, the mutual dependencies between
technology and society are an integral part of the literature on socio-technical systems. More interaction with this field could benefit future AI and data studies \cite{DeWolfEtAl16}.
 So the community finds a stronger reflection of process relevant, but also finds it hard to translate these into concrete research practices. 

Some caveats and encouraging recent developments should be taken into account when interpreting these findings. 

Q1 and Q2 were weakly represented. One reason for this may be that
the requirements on Q1 and Q2 are likely to be less stringent for {\em Social Good} than for the {\em Common Good}. It appears from the definitions put forward by this community (see 
Section \ref{sec-AI4SG}) as well as the conference survey (see Section \ref{sec-summary-of-findings})  that Social Good may well
be produced by considering only or mainly one, potentially very specific, stakeholder group (such as poultry farmers in Africa \cite{poultry} or citizens registering as unemployed in one city \cite{unemployed}). Developing AI for these groups and/or relevant use cases involving them may still require researchers to consider various perspectives and problem versions,
but the scope is much more limited than the ``for all'' (members of a given community) of the Common Good. According to some definitions of the Social Good, it also appears legitimate
to outsource the (social) problem definition to, for example, an NGO or government agency. 

The weak representation of Q1 and Q2 may also be a consequence of research practices.
When researchers depend on the collaboration of a project partner
(for example, to have access to data or to stakeholders), they may face difficulties if they conceptualize the problem in a way that contradicts the project partner's notion. This expectation
may discourage them from exploring other conceptualizations of the problem. 

Regarding the breadth of knowledge brought to the research process (one aspect of Q3), 
a conference not surveyed here made an interesting decision: the ``Fairness, Accountability and Transparency in Machine Learning'' (FATML) workshop organizers decided to host, as of 2018, a conference called FAT* and to turn FATML into one sub-event. This decision contributed to a more multidisciplinary perspective on fairness, accountability and transparency than in the years before, with contributions drawing on a more diverse set of stakeholders and problem formulations. As a member of the Steering Committee of FAT*, I am probably biased to see this conference as a success, but the case shows that this widening of scope is possible and can be highly successful in terms of the number and quality of submissions and attendance rates. 

The method of the conference survey has limitations. 
The coding exercise was, by design, exploratory, and the method simple. In future work, a codebook and more coders will be employed. In addition, the results of the coding exercise also suggest that the widespread absence of the considerations Q1-Q4 in the publications may be (partially) an artifact of publication conventions that favor unambiguity and the appearance of a linear and smooth research process. In addition, many of the surveyed documents were very short and therefore concentrated on telling a simple story; a longer paper may have given room to alternatives considered and other details of the research process. As a result, I expect that qualitative interviews with project participants may yield more information.

\section{Conclusions: towards ethics pen-testing}
This section concludes the article by deriving recommendations for design, and by discussing the scope of the proposals made.
\label{sec-conclusions}

\subsection{Recommendations and ethics pen-testing}

From the previous considerations, some recommendations can be derived for AI researchers and practitioners who want to contribute to the Common Good:

\begin{enumerate}
\item
Study the various faces of knowledge and non-knowledge, be wary of framing 
\item
Identify and involve stakeholders throughout 
\item
Keep looking for individual and common effects; don’t forget the socio-political 
\item
Employ proportionality thinking: suitable \& necessary, balance of interests? 
\item
Focus on whole systems instead of parts (e.g. algorithms) 
\item
Consider feedback loops and causal dynamics 
\item
Draw on the state of the art: e.g. anonymization, discrimination-aware methods 
\item
Ask questions, early and again 
\item
Be ready to be wrong, and embrace learning! 
\end{enumerate}
These recommendations follow from the discussions of the four questions and their interlinkage, i.e. in particular from Q3 (1. and 4.), 
Q1 and Q2 (2.), Q2 (3. and 4.)
Q4 (5. and 6.), and the concept of provocations as such (8. and 9.). I added 7. as a general design principle learned from
the discussions around the GDPR, which, to protect the rights and freedoms of individuals affected by (data processing) technology, requires technology and organizational designers
to design
based on the state of the art (see \cite{SchiffnerEtAl18} for a discussion of the challenge of translating this legal requirement into engineering practice).
This requirement appears pertinent and useful to all design ``for Good''.

But how can we ensure that these questions, which are necessarily somewhat vague and ``social-sciency'', actually get asked, and their answers given attention? 
For this, I propose to draw on an established technique from a core computer-science field, IT security. This method can also be regarded as transposing humanistic self-reflective and self-critical thinking into computer-science mindsets. Specifically, the proposal is inspired by pen-testing in IT security. A penetration test, colloquially known as a pen test, is an authorized simulated attack on a computer system that looks for security weaknesses, potentially gaining access to the system's features and data. It is well-known that no system is 100\% secure. The point is not to pass all pen-tests (this is impossible), the point is to get better through trying. Since IT security researchers and professionals know this, pen-testing is considered valuable and integral to development. Analogous thinking is applied in other subfields of computer science: cryptography and anonymization (where attack modeling is key to any formal proposal), and adversarial machine learning.
I propose to apply this mindset in AI design and call the resulting method ethics pen-testing. This consists of asking the lead questions described above with regard to a proposed design, with the intention of ``attacking'' the good intentions, the claim to a contribution to the Common Good -- as a critical and adversarial method for identifying weaknesses not in order to ``fix'' the design or make it unequivocally good (because this is impossible), but in order to make it better. Such ethics pen-testing should be carried out not by the researcher/developer themselves, but by others, including representatives of various stakeholder groups. Drawing on techniques from various disciplines, we have sketched the organization of such ``attacks'' as ``tool clinics'' \cite{MortonEtAl13}.

\subsection{How specific is this article to AI, and to the goal of the Common Good?}

This article was motivated both by the challenges for AI and the specific current concern within AI about the Good and the Common Good.
The examples and proposed countermeasures draw specifically on AI, often DS, methods and applications.
But is, or in what sense(s) is it specific to AI?

As the discussions of data science and robot(ic)s has shown, it is usually not straightforward to analyze, let alone build a ``purely AI'' system. 
Deployed systems, and thus those with real-world ethical implications, usually involve components from many other fields of computer science,
such as user interfaces, network functionalities, or databases.
The question is then which of these components is ``responsible'' for real-world effects. For example, are the often-discussed filter bubbles,
fake news, and potential manipulation problems a consequence of the AI deployed in social media, or of the way online social networks are connected
and information is spread over the Internet? I have heard AI colleagues protective of the supposedly value-free nature of their subfield 
blame the networking features, and network researchers blame the AI. Conversely, positive effects tend to be attributed to one's ``own'' field.
Any researcher who uses the lens of socio-technical systems studies would of course observe immediately 
that indeed these various components together constitute the IT system that creates the effects, and they would point to the effects of the larger socio-technical systems
in which the IT system is embedded. 

For these reasons, neither the conceptual analysis presented in this paper, nor the normative statements derived from it, nor the intervention sketched as an approach
to operationalizing these statements, can be limited to AI in a clean analytical way. In particular, the problem-solving mindset and the difficulties of stakeholder involvement
(lead questions 1 and 2) are integral to computer scientists' and other engineers' professional mindset and practice. To some extent, all interventions share these problems;
the term ``social engineering'' emphasizes this commonality. I suspect that the reductionist tendencies in conceptualizing problems and solutions are
stronger among computer-science engineers, but this presumption should be investigated in future work.

Extending the same reasoning, one could apply the ``adversarial'' method of ethics pen-testing also in a much
wider range of settings. 
For example, methods for
constructive technology assessment (CTA) \cite{Genus06}
involve multi-perspective criticism and (friendly) ``attacks''. The question how
similar or dissimilar they are to pen-testing should be the subject of future research.

With regard to lead questions 3 and 4, AI is particularly implicated due to its special relationship with knowledge (lead question 3) and 
on account of 
the specific side effects and dynamics (lead question 4) of big data, which is a key driver for AI.

In sum, the issues and recommendations of the present paper are characteristic of, but not limited to, artificial intelligence and data science.

Another question of scope concerns the ethical goal: Are the questions and methods proposed here specific to the Common Good?
I believe that they are applicable for other goals too, but the questions to be asked will depend on the specific technology fields and design goals.
Thus, for example, design in the interest of one specific right or value could also profit from asking the questions and using the recommendation checklist. 
But the scope may be smaller, and conflicting viewpoints may not need to be considered, or considered only as a background given, 
in system design when the focus is on one or a small number of stakeholder groups or goals. The interactive system designed together
with an anti-displacement NGO in \cite{AuerbachEtAl17}, already referred to in the conference survey, 
is a good example of a specific focus in stakeholders, goal, and background.

\subsection{A road ahead}
The idea of ethics pen-testing faces one key challenge: that the testers be perceived as ``nagging'' and the tested researcher/developer feel attacked in their personal good intentions. Both concerns have been voiced towards me often in earlier presentations.
It needs to be kept in mind that in security pen-testing, everybody agrees on what the core goal is and when security is ``broken'',
whereas the reason why one stakeholder considers a system ``broken'' in ethics pen-testing may be a goal that the designer does not even agree with
(which implies that there may be debates on whether the system is broken at all).
These challenges can only be overcome if researchers and practitioners take the ethical quality of their products as seriously as the formal and engineering qualities, and if they regard this as a professional rather than personal virtue. 

Finally, let us not forget that we need to attack the choice of reflective questions and the idea of ethics pen-testing itself too -- theoretically and on the basis of case studies. This is an invitation to the AI community to test and share their experiences! 

\subsubsection*{Acknowledgements}
This article arose from a keynote given at the March 2017 AI in Asia -- AI for Social Good workshop in Tokyo, Japan, and the great discussions we had there. 
I am also indebted to the participants of the December 2017 workshop on Intersectionality and Algorithmic Discrimination at the Lorentz Centre, Leiden, The Netherlands.
I thank Doris Allhutter,
Robert Kahlert, Geoffrey Rockwell, Salvatore Ruggieri and Roger Vergauwen for valuable comments on earlier drafts.

\theendnotes

\end{document}